\documentclass[letterpaper, 10 pt, conference]{ieeeconf}
\IEEEoverridecommandlockouts
\usepackage{dblfloatfix}
\usepackage{algorithm}
\usepackage{algpseudocode}
\usepackage{cite}
\usepackage{amsmath,amssymb,amsfonts}
\usepackage{float}
\usepackage{graphicx}
\usepackage{booktabs}
\usepackage{textcomp}
\usepackage{xcolor}
\usepackage{url}

\makeatletter
\let\NAT@parse\undefined
\makeatother

\usepackage{hyperref}
\definecolor{robored}{HTML}{A32D35}
\definecolor{roboblue}{HTML}{2F618E}
\DeclareUnicodeCharacter{03C0}{$\pi$}

\def\BibTeX{{\rm B\kern-.05em{\sc i\kern-.025em b}\kern-.08em
    T\kern-.1667em\lower.7ex\hbox{E}\kern-.125emX}}

\begin{document}
% \title{RoboSTD: Zero-Shot Single-to-Dual Transfer via Sagittal Mirroring for Bimanual Learning
\title{Robo\kern-0.5pt\textcolor{robored}{S}\kern-0.5pt\textcolor{orange}{T}\kern-0.5pt\textcolor{roboblue}{D}: Zero-Shot \textcolor{robored}{Single}-\textcolor{orange}{to}-\textcolor{roboblue}{Dual} Transfer via Sagittal Mirroring for Bimanual Learning
\thanks{
$^{*}$Equal contribution.
$^{\dagger}$Corresponding authors.%
}
\thanks{
$^{1}$Beijing Academy of Artificial Intelligence,
$^{2}$Beijing University of Posts and Telecommunications.
}
}

\author{
Hongyu Wu$^{*1,2}$ \quad
Chong Liu$^{*1,2}$ \quad
Shaoxuan Xie$^{1}$ \quad
Shaqi Luo$^{1}$ \quad
Hanyu Feng$^{2}$ \\
Shihan Wu$^{\dagger 1}$ \quad
Guocai Yao$^{\dagger 1}$ \quad
}

\maketitle

\begin{abstract}
Bimanual robot manipulation plays an important role in real-world robotic applications, but the high cost of collecting bimanual demonstrations limits the scalable development of bimanual policy learning. Meanwhile, existing single-arm manipulation data are difficult to directly transfer to bimanual systems due to the Single-to-Dual gap in
action spaces, coordination requirements, and arm-side spatial preference.
To address this issue, we propose RoboSTD, a zero-shot Single-to-Dual data construction method for bimanual robot learning. RoboSTD first applies sagittal-plane mirroring to observations,
proprioception, and actions to construct executable contralateral
pseudo-demonstrations. It then uses an LLM to generate structured
spatio-temporal coordination constraints that guide the reconstruction
of pseudo-bimanual supervision across workspace regions and task stages.
We evaluate RoboSTD on the AgileX ALOHA platform in both RoboTwin simulation and real-world settings. Experiments demonstrate the effectiveness of RoboSTD for bimanual policy learning. Without requiring additional bimanual demonstrations, RoboSTD increases the average number of collected cups to $\mathbf{2.11\times}$ that of the single-arm baseline in the cup collection task. \textit{Project page:}
% \texttt{\detokenize{https://anonymous.4open.science/r/RoboSTD/}}
\textcolor{robored}{\url{https://huahungy.github.io/RoboSTD/}}
\end{abstract}

% \begin{IEEEkeywords}
% Bimanual Manipulation, Skill Transfer, Data Construction, Demonstration Reuse, Handedness Bias.
% \end{IEEEkeywords}

\section{Introduction}
Bimanual robot manipulation enables complex tasks that are difficult to efficiently perform with single-arm systems through coordinated use of two arms, making it important for real-world robotic applications~\cite{zhao2023aloha,fu2024mobilealoha}. However, compared with single-arm learning, bimanual policy learning faces a more severe data bottleneck. Collecting high-quality bimanual demonstrations requires not only recording the trajectories of both arms, but also maintaining spatial relationships, execution timing, and task-role coordination, making data acquisition substantially more complex and expensive. Previous studies have identified demonstration collection cost and
teleoperation burden as major bottlenecks for scalable robot learning
~\cite{luo2024hajl,mandlekar2023mimicgen,jiang2024dexmimicgen}. Therefore, reducing the dependence on real bimanual demonstrations is critical for scalable bimanual robot learning.

\begin{figure}[!t]
\centering
\includegraphics[width=\columnwidth,height=0.78\textheight,keepaspectratio]{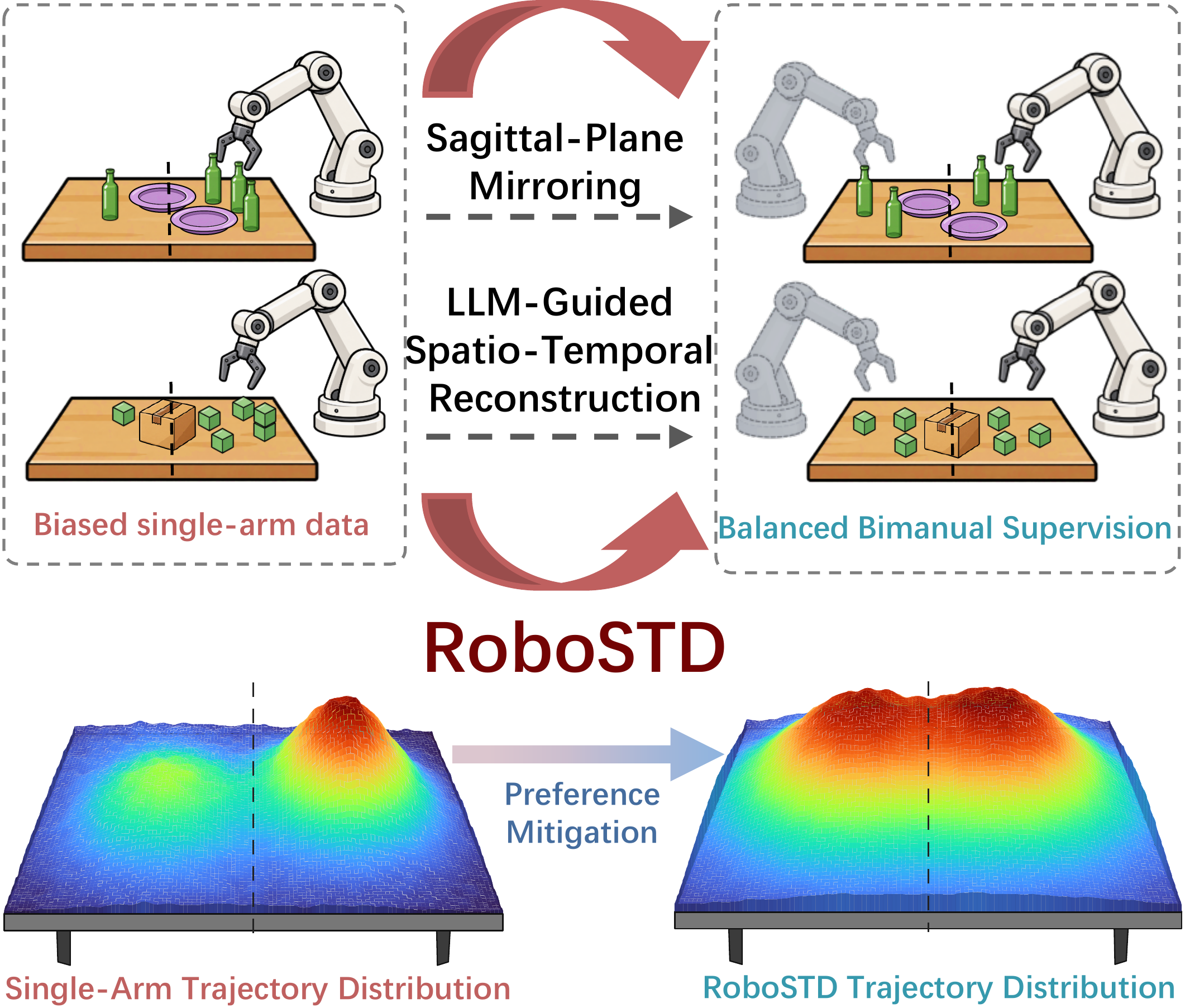}
\caption{\textbf{Overview and motivation of RoboSTD.}
RoboSTD transforms spatially biased single-arm demonstrations into
pseudo-bimanual supervision through two stages: sagittal-plane mirroring
constructs executable contralateral skills, while LLM-guided
spatio-temporal reconstruction coordinates skills through
arm assignment, precedence, and conflict constraints, broadening
workspace coverage and mitigating arm-side spatial preference.}
\label{fig:robostd_overview}
\end{figure}

\begin{figure}[!t]
\centering
\includegraphics[width=\columnwidth,height=0.78\textheight,keepaspectratio]{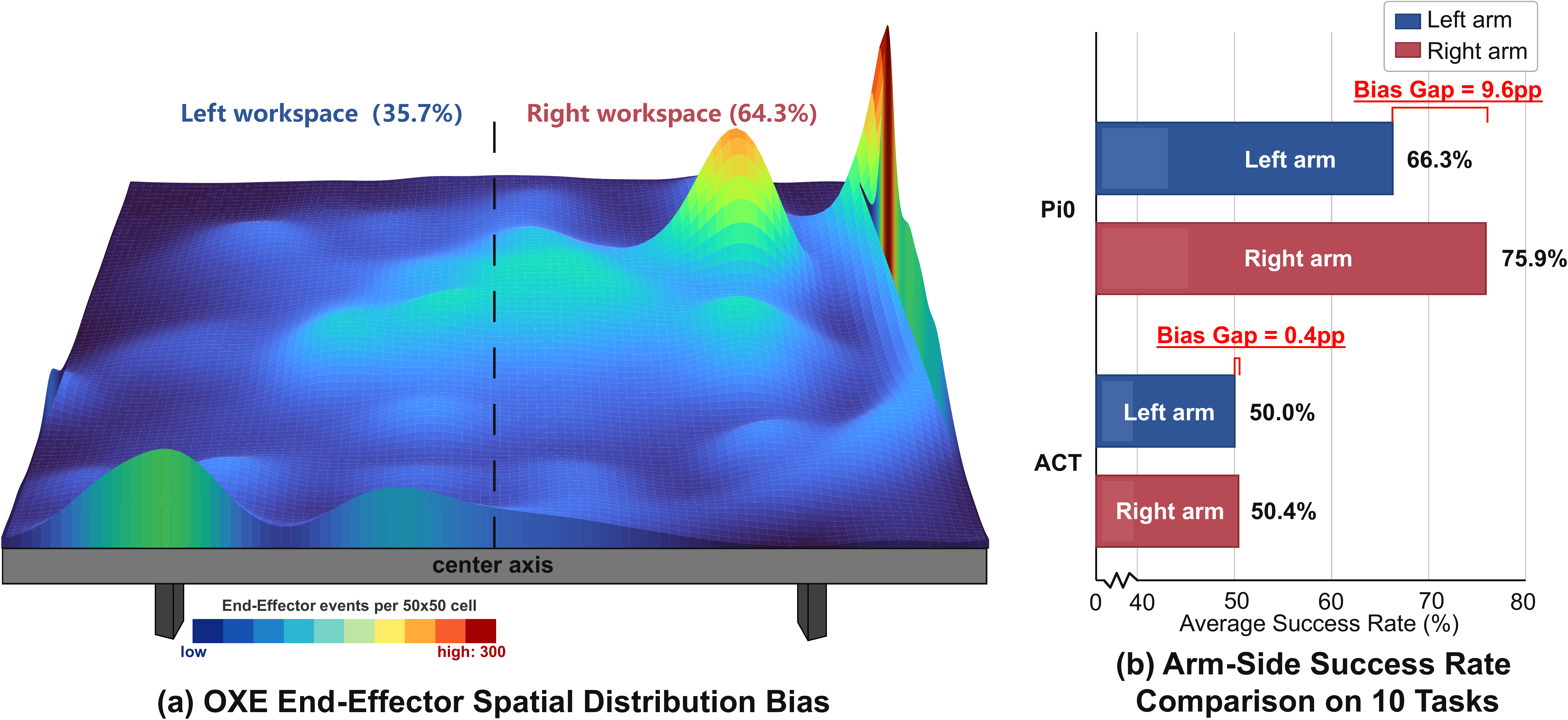}
\caption{\textbf{Arm-side spatial preference analysis in robot data and
policies.}
(a) End-effector workspace distribution estimated from
Open X-Embodiment trajectories to characterize workspace bias.
(b) Left- and right-arm performance comparison of pretrained $\pi_0$
and non-pretrained ACT across 10 RoboTwin 2.0 tasks, examining whether
data-level spatial imbalance is reflected in downstream policy
performance.}
\label{fig:introduction}
\end{figure}

Meanwhile, the amount of single-arm manipulation data accumulated by the robot learning community represents a valuable resource for robot learning. However, how to reuse such single-arm data for bimanual learning remains underexplored. This gives rise to a \textit{Single-to-Dual gap}, involving differences in action spaces, coordination requirements, and spatial preferences across arm sides. An overview of this challenge and our approach is shown in
Fig.~\ref{fig:robostd_overview}. To investigate this phenomenon, we randomly sample 10,000 trajectories
from Open X-Embodiment~\cite{oneill2023openx} and compute the spatial
locations of end-effector events. As shown in
Fig.~\ref{fig:introduction}(a), only \textbf{35.7\%} of these events
occur in the left workspace, whereas \textbf{64.3\%} occur in the right
workspace, corresponding to a \textbf{28.6} percentage-point spatial
imbalance. We further evaluate the pretrained
$\pi_0$~\cite{black2024pi0} and the non-pretrained
ACT~\cite{zhao2023aloha} on 10 tasks in
RoboTwin 2.0~\cite{chen2025robotwin2}. As shown in
Fig.~\ref{fig:introduction}(b), $\pi_0$ achieves average success rates
of \textbf{66.3\%} and \textbf{75.9\%} on left- and right-arm tasks,
respectively, corresponding to a Bias Gap of \textbf{9.6 percentage points}. 
In contrast, ACT achieves \textbf{50.0\%} and \textbf{50.4\%},
with a Bias Gap of only \textbf{0.4 percentage points}.
The right-side preference observed in both the data distribution and
the pretrained policy suggests that upstream spatial imbalance may be
reflected in arm-side performance, potentially hindering
balanced Single-to-Dual transfer.

To address this gap, we propose \textbf{RoboSTD}, a zero-shot
Single-to-Dual data construction method for bimanual robot learning.
Here, \textit{zero-shot} denotes requiring no additional task-specific
bimanual demonstrations, rather than zero-shot policy inference.
RoboSTD consists of two stages. First, \textbf{sagittal-plane mirroring}
transforms observations, proprioception, and actions
~\cite{ordonez2024morphological,zhuang2025mirrorduo} to construct
executable contralateral pseudo-demonstrations. Second, LLM-guided spatio-temporal reconstruction infers arm assignment,
precedence, and conflict constraints from context to compose
original and mirrored skills into pseudo-bimanual supervision.

We evaluate RoboSTD using the AgileX ALOHA platform in both the RoboTwin 2.0
simulation environment~\cite{chen2025robotwin2} and real-world settings.
In the simulation effectiveness experiments over 10 RoboTwin tasks, policies
trained with 100 native bimanual demonstrations per task achieve an
average success rate of \textbf{67.6\%}, while those trained with
RoboSTD-generated supervision from only 50 single-arm demonstrations
per task achieve \textbf{66.5\%}. The small \textbf{1.1} percentage-point
gap shows that RoboSTD can construct effective bimanual supervision from fewer single-arm demonstrations. In the spatial generalization task, RoboSTD extends skills originally limited to a unilateral workspace to the full bimanual workspace, achieving a $\mathbf{2.11\times}$ increase in the average number of collected cups over the single-arm baseline. In the long-horizon sandwich-making task, RoboSTD further improves the
task success rate and average completion, demonstrating stronger temporal coordination. These results demonstrate that sagittal-plane mirroring and
LLM-guided spatio-temporal reconstruction can effectively extend
single-arm demonstrations into useful bimanual supervision.

The main contributions are summarized as follows:
\begin{itemize}
    \item We identify and analyze arm-side spatial preferences in robot data and pretrained policies, and reveal their potential impact on balanced bimanual learning.
    \item We propose RoboSTD, a zero-shot Single-to-Dual data construction method that converts existing single-arm demonstrations into bimanual supervision.
    \item We demonstrate RoboSTD's effectiveness for Single-to-Dual transfer, improving workspace coverage and long-horizon coordination in real-world bimanual tasks.
\end{itemize}

\section{Related Work}
\begin{table*}[!t]
\caption{Comparison of Data Reuse and Single-to-Dual Transfer Methods for Bimanual Robot Learning}
\label{tab:related_work_comparison}
\centering
\small
\setlength{\tabcolsep}{4pt}
\renewcommand{\arraystretch}{1.05}

\resizebox{\textwidth}{!}{%
\begin{tabular}{lcccccc}
\toprule
\textbf{Method} &
\textbf{\shortstack[c]{Source\\Supervision}} &
\textbf{\shortstack[c]{No Additional\\Bimanual}} &
\textbf{\shortstack[c]{Automated Data\\Generation}} &
\textbf{\shortstack[c]{Sagittal\\Mirroring}} &
\textbf{\shortstack[c]{Single-to-\\Dual}} &
\textbf{\shortstack[c]{Spatio-Temporal\\Supervision}} \\
\midrule

DexMimicGen &
Bimanual demos &
$\times$ &
$\checkmark$ &
$\times$ &
$\times$ &
$\circ$ \\

D-CODA &
Bimanual demos &
$\times$ &
$\checkmark$ &
$\times$ &
$\times$ &
$\times$ \\

ROPA &
Bimanual demos &
$\times$ &
$\checkmark$ &
$\times$ &
$\times$ &
$\times$ \\

AnyBimanual &
Pretrained unimanual policy &
$\times$ &
$\times$ &
$\times$ &
$\checkmark$ &
$\circ$ \\

\textbf{RoboSTD} &
\textbf{Single-arm demos} &
$\checkmark$ &
$\checkmark$ &
$\checkmark$ &
$\checkmark$ &
$\checkmark$ \\

\bottomrule
\end{tabular}
}

\vspace{1pt}
\begin{minipage}{\textwidth}
\footnotesize
\textit{Note:}
$\checkmark$: explicitly supported;
$\circ$: partially or implicitly supported;
$\times$: not supported.
\end{minipage}

\end{table*}

\textbf{Unimanual and Bimanual Robot Datasets.}
Large-scale robot datasets have become an important foundation for learning generalizable manipulation policies. Open X-Embodiment~\cite{oneill2023openx} aggregates large-scale
cross-embodiment robot data, while DROID~\cite{khazatsky2024droid}
and BridgeData~\cite{walke2023bridgedata} provide diverse real-world
manipulation data across tasks and environments. However, most existing
large-scale resources are dominated by unimanual interaction. For bimanual manipulation, ALOHA~\cite{zhao2023aloha} provides a low-cost dual-arm teleoperation platform for collecting coordinated demonstrations, while Mobile ALOHA~\cite{fu2024mobilealoha} extends data collection to mobile bimanual manipulation. RoboCOIN~\cite{wu2025robocoin} further provides real-world bimanual manipulation data for learning and evaluating coordinated policies. In simulation, RoboTwin 2.0~\cite{chen2025robotwin2} provides scalable bimanual task environments and data-generation pipelines. Despite these efforts, large-scale, high-quality real-world bimanual data remain substantially more difficult and expensive to acquire than unimanual data.

\textbf{Unimanual and Bimanual Policy Learning.}
Recent imitation-learning methods have substantially improved robot
manipulation through sequence modeling~\cite{zhao2023aloha},
diffusion-based action generation~\cite{chi2023diffusion}, and
large-scale generalist or vision-language-action pretraining
~\cite{black2024pi0,kim2024openvla,ghosh2024octo}. Compared with unimanual manipulation, bimanual policy learning must jointly model the actions of both arms together with their spatial and temporal coordination, introducing additional challenges in role assignment and long-horizon execution. ALOHA~\cite{zhao2023aloha} and Mobile ALOHA~\cite{fu2024mobilealoha}
demonstrate bimanual policy learning from coordinated demonstrations.
These methods highlight the importance of jointly modeling both arms
together with their spatial and temporal coordination in complex
bimanual tasks.

\textbf{Robot Data Augmentation and Bimanual Transfer.}
Another line of work improves data efficiency by augmenting or reusing
robot demonstrations. MimicGen~\cite{mandlekar2023mimicgen} and
DexMimicGen~\cite{jiang2024dexmimicgen} automatically generate
demonstrations from a small number of source demonstrations, while
D-CODA~\cite{liu2025dcoda} and ROPA~\cite{chen2025ropa} improve the
utilization of existing bimanual demonstrations through offline data
augmentation. AnyBimanual~\cite{lu2024anybimanual} transfers pretrained
unimanual policies to bimanual manipulation, but still relies on
additional bimanual demonstrations for adaptation. Other studies exploit
morphological symmetry to improve sample efficiency and bilateral
generalization~\cite{ordonez2024morphological,li2025symdex}.
As summarized in Table~\ref{tab:related_work_comparison}, existing
approaches differ in their source supervision, dependence on additional
bimanual demonstrations, and ability to support Single-to-Dual transfer
and spatio-temporal coordination.

\section{Problem Formulation}
We study zero-shot Single-to-Dual transfer: constructing bimanual
supervision from existing single-arm demonstrations without collecting
additional task-specific bimanual demonstrations. Here, zero-shot refers
to the data requirement rather than policy inference. For arm $k\in\{L,R\}$, let
\begin{equation}
\tau^k
=
\{(o_t,s_t^k,a_t^k)\}_{t=1}^{T},
\end{equation}
and our goal is to transform single-arm demonstrations into executable
coordinated supervision for a bimanual policy.

A single-arm trajectory observes only one side of the interaction,
whereas bimanual supervision requires the states and actions of both
arms together with spatio-temporal coordination. Given only $\tau^k$,
the contralateral state $s_t^{\bar{k}}$, action $a_t^{\bar{k}}$, and
coordination annotation $c_t$ are unavailable, where $c_t$ denotes the
language-form annotation of the current subtask, arm role, and temporal
stage. The action space expands from
\begin{equation}
\mathcal{A}_{\mathrm{single}}=\mathcal{A}^{k},
\qquad
\mathcal{A}_{\mathrm{bi}}
=
\mathcal{A}^{L}\times\mathcal{A}^{R},
\end{equation}
making direct reuse of single-arm demonstrations generally insufficient for
coordinated bimanual policy learning.

Recovering valid bimanual supervision from single-arm data requires
more than flipping observations or copying robot commands. Differences
in coordinate frames, joint axes, and mounting configurations can break
the correspondence between observations, proprioceptive states, and
actions across the two arms. Therefore, Single-to-Dual transfer requires a cross-arm mapping
that preserves consistency.

Even with valid mirrored skills, directly combining original and
mirrored trajectories does not guarantee coordinated bimanual behavior.
The two arms may enter overlapping workspace regions or execute
incompatible actions at the same stage. Long-horizon tasks require
arm assignment and sequencing to avoid spatial conflicts and
temporal ambiguity.

Given only $\mathcal{D}_{\mathrm{single}}$, Single-to-Dual aims to
construct
\begin{equation}
\widetilde{\mathcal{D}}_{\mathrm{bi}}
=
\mathcal{F}
\left(
\mathcal{D}_{\mathrm{single}}
\right),
\end{equation}
such that the resulting supervision is physically executable and
spatio-temporally coordinated for bimanual policy learning. This leads
to two core motivations:

\begin{center}
\fbox{
\begin{minipage}{0.92\linewidth}
\textbf{Motivation 1: Physically Valid Cross-Arm Transfer.}
How can single-arm demonstrations be mapped across arms with
geometric and control consistency?

\vspace{2pt}

\textbf{Motivation 2: Coordinated Spatio-Temporal Composition.}
How can transferred skills be organized without workspace conflicts or
temporal ambiguity?
\end{minipage}
}
\end{center}
\begin{figure*}[!t]
\centering
\includegraphics[width=\textwidth]{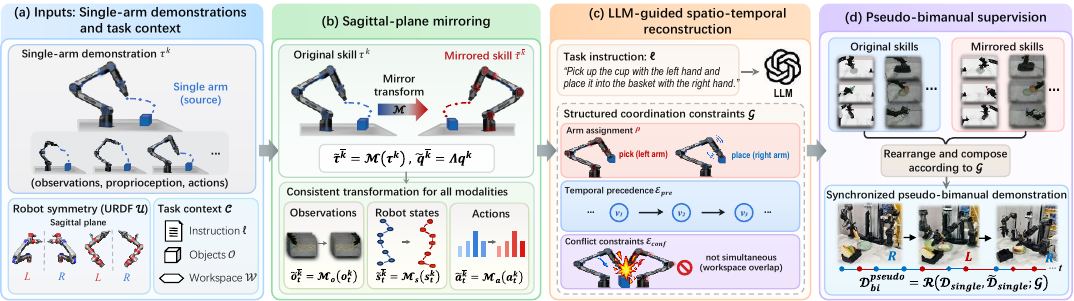}
\caption{\textbf{Overview of RoboSTD.}
(a) Inputs include single-arm demonstrations, robot geometry, and task
context. (b) Sagittal-plane mirroring jointly transforms observations,
proprioception, and actions to generate contralateral
pseudo-demonstrations. (c) An LLM infers arm assignments, precedence,
and conflict constraints. (d) These constraints guide the composition
of original and mirrored skills into pseudo-bimanual supervision.}
\label{fig:method_overview}
\end{figure*}
\section{Method}
RoboSTD constructs pseudo-bimanual supervision from single-arm
demonstrations, robot geometry, and task context, as shown in
Fig.~\ref{fig:method_overview}(a). Let
$\mathcal{D}_{\mathrm{single}}$ denote the source dataset,
$\mathcal{U}$ the robot URDF (Unified Robot Description Format), and
$\mathcal{C}=(\ell,\mathcal{O},\mathcal{W})$ the task context.
The two stages, sagittal-plane mirroring and LLM-guided spatio-temporal
reconstruction, jointly enable skill transfer and coordination without
collecting additional task-specific bimanual demonstrations
(Fig.~\ref{fig:method_overview}(b--d)).
Algorithm~\ref{alg:robostd} summarizes the pipeline.

\subsection{Sagittal-Plane Mirroring}
\label{sec:sagittal_mirroring}

Motivated by morphological symmetry~\cite{ordonez2024morphological}
and multimodal reflection~\cite{zhuang2025mirrorduo}, RoboSTD constructs
contralateral supervision using a platform-specific mapping.
Given a trajectory
$\tau^{k}=\{(o_t^{k},s_t^{k},a_t^{k})\}_{t=1}^{T}$ from arm
$k\in\{L,R\}$, $o_t^k$, $s_t^k$, and $a_t^k$ denote the visual
observation, proprioceptive state, and action at timestep $t$,
respectively, and $T$ is the trajectory length.
The pseudo-demonstration for the opposite arm $\bar{k}$ is defined
\begin{equation}
\tilde{\tau}^{\bar{k}}
=
\mathcal{M}(\tau^{k})
=
\left\{
(\tilde{o}_t^{\bar{k}},
\tilde{s}_t^{\bar{k}},
\tilde{a}_t^{\bar{k}})
\right\}_{t=1}^{T},
\label{eq:robostd_mirrored_trajectory}
\end{equation}
where a tilde denotes a mirrored quantity.
The operator $\mathcal{M}$ exchanges arm identity and applies
$\tilde{o}_t^{\bar{k}}=\mathcal{M}_o(o_t^k)$,
$\tilde{s}_t^{\bar{k}}=\mathcal{M}_s(s_t^k)$, and
$\tilde{a}_t^{\bar{k}}=\mathcal{M}_a(a_t^k)$.
Here, $\mathcal{M}_o$, $\mathcal{M}_s$, and $\mathcal{M}_a$ are the
observation, state, and action mappings for sagittal
reflection, preserving their physical correspondence under the
symmetry. Applying $\mathcal{M}$ to $\mathcal{D}_{\rm single}$ yields
$\tilde{\mathcal{D}}_{\rm single}$.

In a robot-centered frame with its origin on the sagittal plane, the
reflection matrix $\mathbf{S}$ satisfies
$\mathbf{S}=\mathbf{S}^{T}=\mathbf{S}^{-1}$ and
$\det(\mathbf{S})=-1$.
Let $\mathbf{p}$ and $\mathbf{R}$ denote the end-effector position
and orientation matrix. With matching world- and end-effector-frame
reflection conventions, their mirrored counterparts are
$\tilde{\mathbf{p}}=\mathbf{S}\mathbf{p}$ and
$\tilde{\mathbf{R}}=\mathbf{S}\mathbf{R}\mathbf{S}$.
Writing $\mathbf{R}(\mathbf{v},\theta)$ for a rotation by angle
$\theta$ about unit rotation axis $\mathbf{v}$ gives
\begin{equation}
\mathbf{S}\mathbf{R}(\mathbf{v},\theta)\mathbf{S}
=
\mathbf{R}(\mathbf{S}\mathbf{v},-\theta),
\label{eq:robostd_axis_angle_reflection}
\end{equation}
which determines the corresponding rotation-axis and angle-sign changes required for a physically mirrored orientation.

For kinematically symmetric arms with matched joint ordering and
mirror-compatible zero configurations, we use
\begin{equation}
\tilde{\mathbf{q}}^{\bar{k}}
=
\mathbf{\Lambda}\mathbf{q}^{k},
\qquad
\mathbf{\Lambda}
=
\mathrm{diag}(\lambda_1,\ldots,\lambda_n),
\label{eq:robostd_joint_mapping}
\end{equation}
where $\mathbf{q}^{k}\in\mathbb{R}^{n}$ contains the $n$ revolute-joint
angles of arm $k$, and $\lambda_i\in\{+1,-1\}$ determines whether
joint $i$ preserves or reverses its angle sign. Let
$\mathbf{v}_i^k$ and $\mathbf{v}_i^{\bar{k}}$ denote corresponding
unit joint axes at reference configurations in a
common frame. The joint signs follow from the reflected-axis relation:
\begin{equation}
\lambda_i =
\begin{cases}
+1, & \mathbf{v}_i^{\bar{k}} = -\mathbf{S}\mathbf{v}_i^k,\\
-1, & \mathbf{v}_i^{\bar{k}} = \mathbf{S}\mathbf{v}_i^k.
\end{cases}
\label{eq:joint_sign}
\end{equation}
A detailed derivation of the joint-space mapping, including the
physically mirrored and isomorphic-arm cases, is provided in
Appendix~\ref{app:joint_mapping}.
The special case $\mathbf{\Lambda}=-\mathbf{I}$, with $\mathbf{I}$ the
$n$-dimensional identity matrix, yields
$\tilde{\mathbf{q}}^{\bar{k}}=-\mathbf{q}^{k}$. RoboSTD derives
$\mathbf{\Lambda}$ from joint-axis definitions and mounting configuration
and applies it consistently to the corresponding joint states and actions.

\subsection{LLM-Guided Spatio-Temporal Reconstruction}
\label{sec:spatiotemporal}

Sagittal-plane mirroring provides executable skills for both arms, but
independent single-arm trajectories do not specify how these skills
should be assigned and coordinated in a bimanual task. RoboSTD therefore
uses an LLM to infer structured spatio-temporal constraints from task
context $\mathcal{C}=(\ell,\mathcal{O},\mathcal{W})$.
Here, $\ell$ is the language instruction, $\mathcal{O}$ contains object
identities and locations, and $\mathcal{W}$ specifies workspace,
reachability, and safety constraints for both arms.

We first decompose the task into $N$ manipulation units
$\mathcal{V}=\{v_i\}_{i=1}^{N}$, where each $v_i$ denotes a segment
associated with an available original or mirrored skill. Given $\mathcal{V}$ and task context
$\mathcal{C}=(\ell,\mathcal{O},\mathcal{W})$, the LLM generates
the following structured coordination constraints, as shown in
Fig.~\ref{fig:method_overview}(c):
\begin{equation}
\mathcal{G}
=
\mathrm{LLM}
\left(
\mathcal{V},\ell,\mathcal{O},\mathcal{W}
\right)
=
\left(
\boldsymbol{\rho},
\mathcal{E}_{\mathrm{pre}},
\mathcal{E}_{\mathrm{conf}}
\right),
\end{equation}
where $\boldsymbol{\rho}=\{\rho_i\}_{i=1}^{N}$ contains the arm
assignments, with $\rho_i\in\{L,R\}$ identifying the arm responsible
for unit $v_i$. The directed relation
$(v_i,v_j)\in\mathcal{E}_{\mathrm{pre}}$ requires $v_i$ to precede
$v_j$, while $\mathcal{E}_{\mathrm{conf}}$ contains pairs of units that
should not execute simultaneously due to workspace overlap or task
conflicts.

In implementation, we use GPT-4.1~\cite{openai2025gpt41} as the
planner. It returns $\boldsymbol{\rho}$,
$\mathcal{E}_{\mathrm{pre}}$, and $\mathcal{E}_{\mathrm{conf}}$ via
a predefined JSON schema, which is automatically validated before
reconstruction.

As shown in Fig.~\ref{fig:method_overview}(d), the rearrangement
operator $\mathcal{R}$ uses these constraints to compose original and
mirrored demonstrations into the pseudo-bimanual dataset:
\begin{equation}
\mathcal{D}_{\mathrm{bi}}^{\mathrm{pseudo}}
=
\mathcal{R}
\left(
\mathcal{D}_{\mathrm{single}},
\tilde{\mathcal{D}}_{\mathrm{single}};
\mathcal{G}
\right).
\end{equation}
For each unit, $\mathcal{R}$ selects the original or mirrored skill
according to $\rho_i$, orders the selected skills using
$\mathcal{E}_{\mathrm{pre}}$, and temporally separates conflicting units
in $\mathcal{E}_{\mathrm{conf}}$. During a single-arm stage, the inactive
arm maintains its current configuration. Both arm sequences are then
aligned on a common timeline, yielding pseudo-bimanual samples
$(o_t,s_t^L,s_t^R,a_t^L,a_t^R)$.

Each reconstructed stage is further assigned a language-form annotation
$c_t$ encoding its subtask, arm role, and temporal stage. We augment the
task instruction as $\ell_t^{\mathrm{coord}}=\ell\oplus c_t$, where
$\oplus$ denotes textual composition. The resulting supervision trains
the language-conditioned bimanual policy
\begin{equation}
\pi
\left(
a_t^L,a_t^R
\mid
o_t,s_t,\ell_t^{\mathrm{coord}}
\right),
\end{equation}
where $s_t=(s_t^L,s_t^R)$ is the combined proprioceptive state.
RoboSTD therefore modifies the training supervision without changing
the downstream policy architecture.

\subsection{Algorithm Summary}
\begin{algorithm}[H]
\caption{RoboSTD Data Construction}
\label{alg:robostd}
\small
\algrenewcommand{\algorithmicindent}{1.2em}
\begin{algorithmic}[1]

\Require Single-arm dataset $\mathcal{D}_{\mathrm{single}}$;
robot URDF $\mathcal{U}$ and mounting configuration;
task context $\mathcal{C}=(\ell,\mathcal{O},\mathcal{W})$

\Ensure Pseudo-bimanual dataset
$\mathcal{D}_{\mathrm{bi}}^{\mathrm{pseudo}}$

\State Determine $\mathbf{S}$ from robot geometry and mounting

\State Derive $\mathbf{\Lambda}$ from common-frame joint axes using
Eq.~\ref{eq:joint_sign}

\State Initialize
$\tilde{\mathcal{D}}_{\mathrm{single}}\gets\varnothing$

\For{each $\tau^k\in\mathcal{D}_{\mathrm{single}}$}
    \State $\tilde{\tau}^{\bar{k}}
    \gets\mathcal{M}(\tau^k;\mathbf{S},\mathbf{\Lambda})$
    \State $\tilde{\mathcal{D}}_{\mathrm{single}}
    \gets\tilde{\mathcal{D}}_{\mathrm{single}}
    \cup\{\tilde{\tau}^{\bar{k}}\}$
\EndFor

\State Decompose task into units
$\mathcal{V}=\{v_i\}_{i=1}^{N}$

\State Generate coordination constraints
$\mathcal{G}\gets\mathrm{LLM}(\mathcal{V},\mathcal{C})$
\Statex \hspace{\algorithmicindent}
$\mathcal{G}
=
(\boldsymbol{\rho},
\mathcal{E}_{\mathrm{pre}},
\mathcal{E}_{\mathrm{conf}})$

\State Rearrange skills using $\mathcal{G}$ and hold inactive arm poses

\State Construct
$\mathcal{D}_{\mathrm{bi}}^{\mathrm{pseudo}}
\gets
\mathcal{R}
(\mathcal{D}_{\mathrm{single}},
\tilde{\mathcal{D}}_{\mathrm{single}};\mathcal{G})$

\State Add stage annotation $c_t$ to the language condition $\ell$

\State \Return $\mathcal{D}_{\mathrm{bi}}^{\mathrm{pseudo}}$

\end{algorithmic}
\end{algorithm}

\section{Experiments}
% ================================================================
% Experiments
% ================================================================

\begin{figure*}[!t]
\centering
\includegraphics[width=\textwidth,keepaspectratio]
{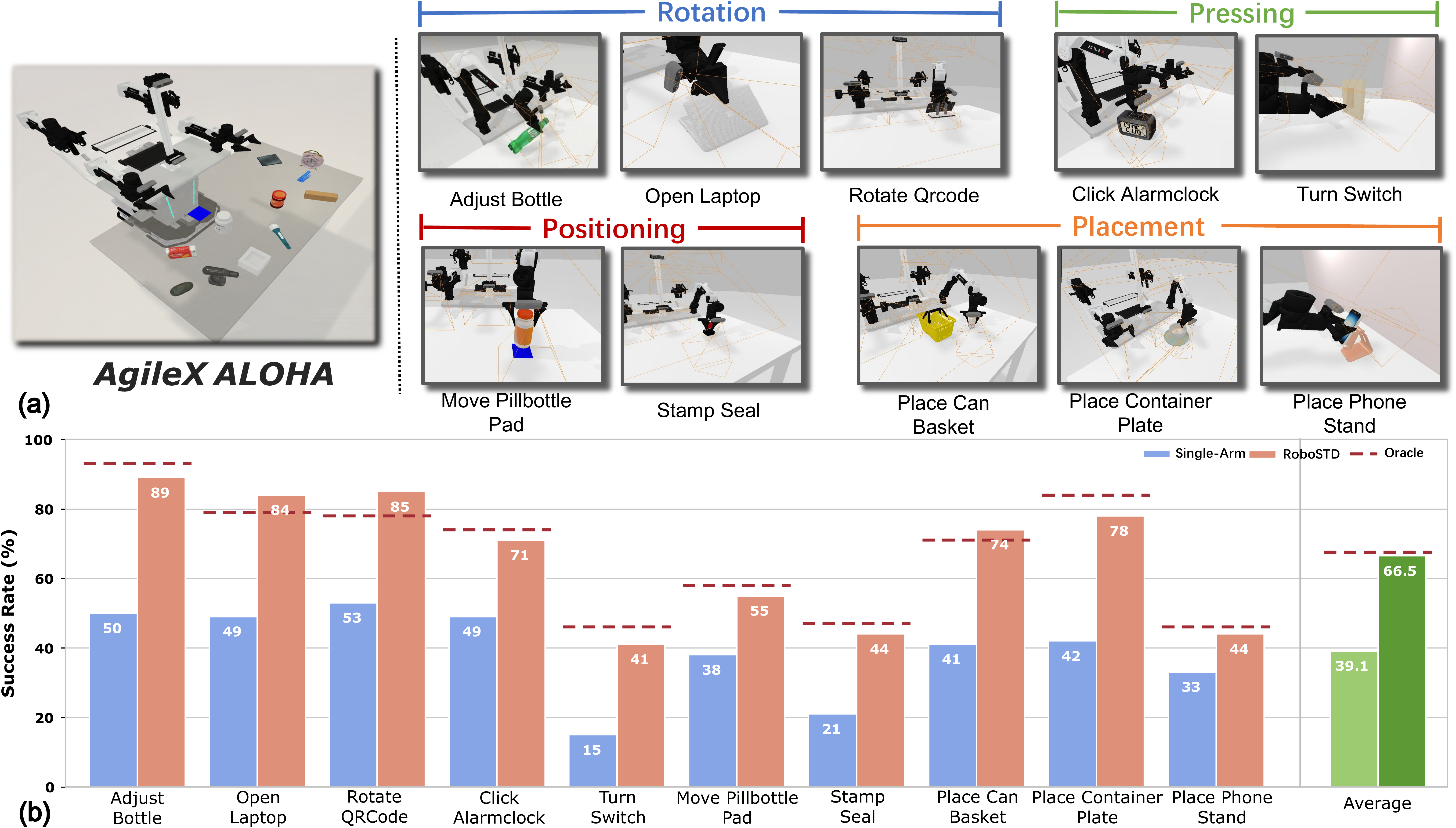}
\caption{\textbf{RoboSTD evaluation on 10 RoboTwin tasks.}
(a) Dual-arm AgileX ALOHA platform and 10 manipulation tasks covering
rotation, pressing, positioning, and placement.
(b) Task-wise success rates of $\pi_0$ under Single-Arm, RoboSTD, and
Oracle supervision settings.}
\label{fig:experiment1_v2}
\end{figure*}

In this section, we comprehensively evaluate RoboSTD in simulation
and real-world environments to answer research questions on
effectiveness, coordination, and bias mitigation:

\begin{center}
\setlength{\fboxsep}{6pt}
\fbox{%
\begin{minipage}{0.95\columnwidth}
\textbf{\textit{Q1.}}
\textit{Can RoboSTD-generated pseudo-bimanual supervision effectively
support bimanual policy learning?}

\textbf{\textit{Q2.}}
\textit{Can LLM-guided spatio-temporal reconstruction improve bimanual coordination?}

\textbf{\textit{Q3.}}
\textit{Can RoboSTD mitigate arm-side spatial preference during
fine-tuning of pretrained robot policies?}
\end{minipage}%
}
\end{center}

% ================================================================
\subsection{Experimental Setup}

We evaluate RoboSTD in simulation and real-world environments using the
dual-arm AgileX ALOHA platform~\cite{zhao2023aloha}.

\textbf{Simulation Environment.}
Simulation experiments are conducted in RoboTwin 2.0~\cite{chen2025robotwin2}
using an AgileX ALOHA robot with two 6-DoF arms. We select 10
manipulation tasks covering rotation, pressing, positioning, and
placement. For the effectiveness study, the Single-Arm and RoboSTD
settings use the same 50 source single-arm demonstrations per task,
while the Oracle setting uses 100 directly collected native bimanual
demonstrations per task. For the arm-side spatial preference study, we use a separate set of
1,000 original single-arm demonstrations, with 100 demonstrations for
each of the 10 RoboTwin tasks, and evaluate three tasks with the largest right-to-left gaps in the preliminary
analysis.

\textbf{Real-World Environment.}
We evaluate five real-world bimanual tasks on the same AgileX ALOHA
platform. Three low-coordination tasks, \textit{Bowl Placement},
\textit{Towel Storage}, and \textit{Flower Arrangement}, evaluate
rigid-object placement, deformable-object manipulation, and precision
placement, respectively. Two high-coordination tasks,
\textit{Cup Collection} and \textit{Sandwich Making}, further evaluate
spatial coordination and long-horizon temporal coordination. For each task, we collect
50 single-arm demonstrations for RoboSTD and 50
native bimanual demonstrations for the Oracle reference.

\textbf{Data Configurations.}
We compare three supervision settings where applicable.
\textit{Single-Arm} directly uses the original source single-arm
demonstrations; \textit{RoboSTD} constructs pseudo-bimanual supervision
from the same source demonstrations; and \textit{Oracle} uses directly
collected native bimanual demonstrations as reference supervision.
For the high-coordination real-world tasks, we additionally evaluate
\textit{RoboSTD w/o LLM}, which retains sagittal-plane mirroring and
spatial rearrangement but removes the LLM-generated arm assignment,
precedence, and conflict constraints. Original and mirrored skills are
composed using the default execution order without explicit
spatio-temporal coordination.

\textbf{Policies.}
We use $\pi_0$~\cite{black2024pi0} for the RoboTwin experiments and
$\pi_{0.5}$~\cite{black2025pi05} for all real-world experiments in this study.
For high-coordination tasks, $c_t$ is incorporated into the language
condition, leveraging $\pi_{0.5}$'s semantic subtask conditioning.

\textbf{Training Details.}
All policies are fine-tuned with a batch size of \textbf{32}. We use
a \textbf{1,000}-step warmup followed by a learning-rate decay schedule
from $2.5\times10^{-5}$ to $2.5\times10^{-6}$ with
\textbf{30,000} decay steps. Training is performed on an NVIDIA A800 GPU
with \textbf{80 GB} memory. All configurations use identical optimization settings to ensure fairness.

\textbf{Evaluation Metrics.}
Task success rate (SR) is reported for all RoboTwin tasks and applicable
real-world tasks. For \textit{Cup Collection}, we additionally report
the average number of collected cups. For \textit{Sandwich Making}, we
report task success rate and average completion level. Arm-side performance imbalance is measured by
$\mathrm{Bias\ Gap}=SR_{\mathrm{right}}-SR_{\mathrm{left}}$, where a
positive value indicates a right-side performance advantage and a
smaller gap indicates more balanced arm-side performance. To quantify
execution-level spatial preference, we project end-effector trajectories
onto the $\Delta X$--$\Delta Z$ workspace plane and define Density Gap as follows:
\begin{equation}
\mathrm{Density\ Gap}
=
\frac{\bar{\rho}_{R}-\bar{\rho}_{L}}
{\bar{\rho}_{L}}\times100\%,
\end{equation}
where $\bar{\rho}_{R}$ and $\bar{\rho}_{L}$ denote the mean trajectory
densities in the right and left workspaces, respectively. A smaller
positive value indicates more balanced left--right workspace coverage.

% ================================================================
\subsection{Effectiveness of RoboSTD for Bimanual Learning}

To answer \textbf{Q1}, we evaluate whether pseudo-bimanual supervision
constructed by RoboSTD can effectively support bimanual policy learning
on 10 manipulation tasks from RoboTwin 2.0~\cite{chen2025robotwin2}. As shown
in Fig.~\ref{fig:experiment1_v2}(a), the selected tasks cover four
representative manipulation categories, including rotation, pressing,
positioning, and placement, providing diverse object interactions and
workspace configurations for evaluating Single-to-Dual transfer. The
Single-Arm and RoboSTD settings use the same 50 source single-arm
demonstrations per task, whereas the Oracle reference uses 100 directly
collected native bimanual demonstrations per task.

As shown in Fig.~\ref{fig:experiment1_v2}(b), the Single-Arm setting
achieves an average success rate of only \textbf{39.1\%}, reflecting
the difficulty of directly reusing single-arm demonstrations for
bimanual execution. In contrast, RoboSTD raises the average success rate
to \textbf{66.5\%}, closely approaching the \textbf{67.6\%} achieved
by the Oracle setting. The difference between RoboSTD and Oracle is only
\textbf{1.1 percentage points (pp)}, despite RoboSTD using 50
single-arm source demonstrations instead of 100 directly collected
bimanual demonstrations. The close performance to Oracle indicates that RoboSTD-generated
pseudo-bimanual supervision provides effective training signals
comparable to native bimanual demonstrations.

The task-wise results in Fig.~\ref{fig:experiment1_v2}(b) further show
that RoboSTD substantially outperforms the Single-Arm setting across all
10 tasks while remaining close to the Oracle reference overall. On
several tasks, including \textit{Open Laptop}, \textit{Place Can Basket},
and \textit{Rotate QRCode}, RoboSTD even achieves higher success rates
than the native bimanual reference. These results suggest
that sagittal-plane mirroring and reconstruction retain the
visual, proprioceptive, and action correspondences required for
Single-to-Dual transfer. Overall, RoboSTD provides effective and
data-efficient bimanual supervision without requiring additional
task-specific bimanual demonstrations for policy training.

% ================================================================
\begin{figure}[!t]
\centering
\includegraphics[width=\columnwidth,keepaspectratio]
{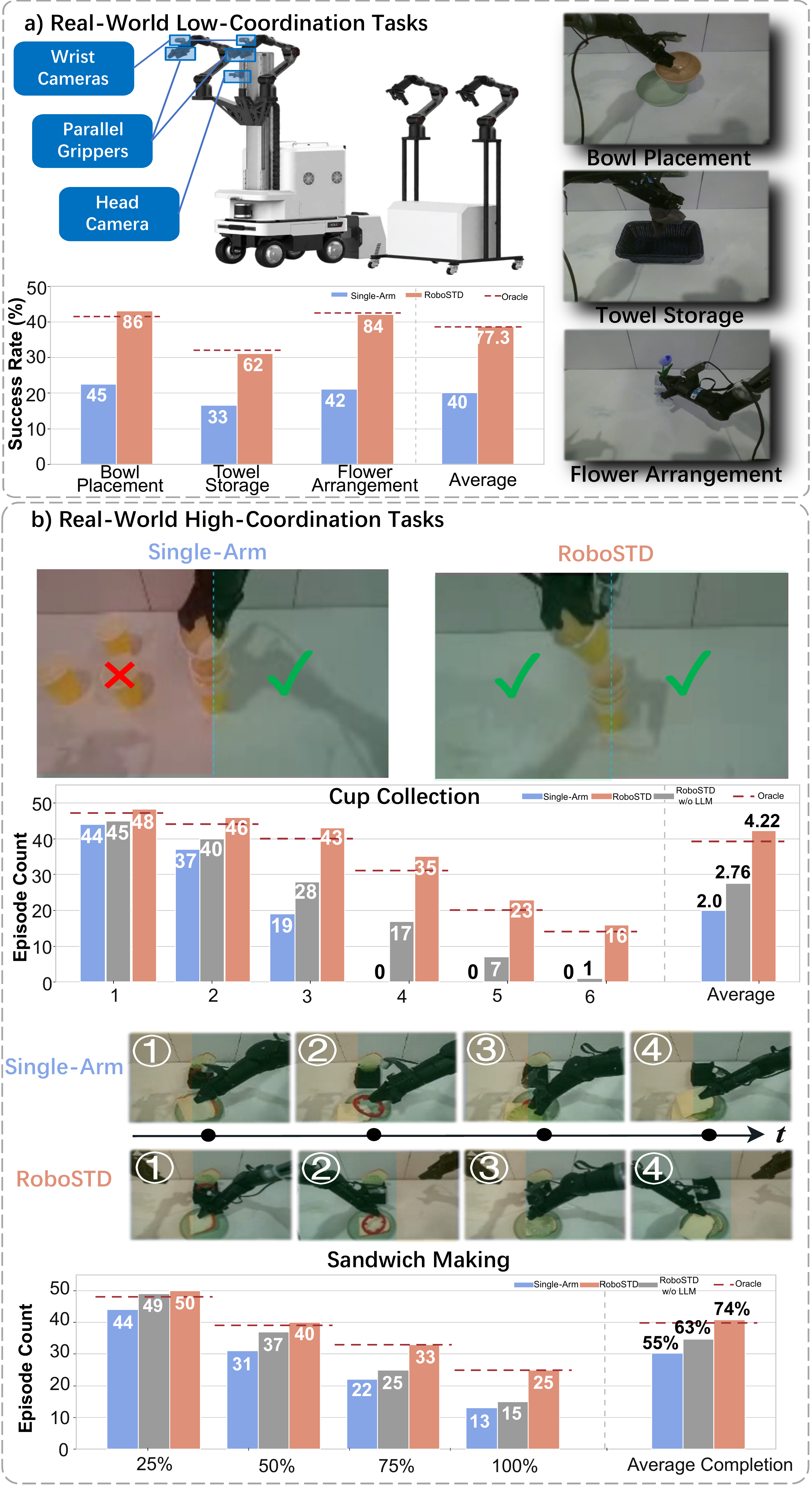}
\caption{\textbf{Real-world evaluation of RoboSTD across coordination levels.}
(a) Low-coordination tasks covering rigid, deformable, and precision manipulation.
(b) High-coordination tasks comparing Single-Arm, RoboSTD w/o LLM,
RoboSTD, and Oracle to evaluate structured spatio-temporal coordination using task-specific metrics.}
\label{fig:experiment2}
\end{figure}

\subsection{Bimanual Manipulation Across Coordination Levels}

We further evaluate RoboSTD on five real-world bimanual tasks using the
dual-arm AgileX ALOHA platform~\cite{zhao2023aloha} and
$\pi_{0.5}$~\cite{black2025pi05}. As shown in
Fig.~\ref{fig:experiment2}, the tasks are organized according to their
inter-arm coordination requirements. The three low-coordination tasks
primarily examine whether single-arm skills can be effectively
transferred into real-world bimanual supervision, while the two
high-coordination tasks further evaluate whether the transferred skills
can be spatially and temporally coordinated. For the latter, \textit{RoboSTD w/o LLM} retains the geometric
Single-to-Dual construction while removing explicit LLM-generated
coordination constraints, isolating the contribution of the
spatio-temporal reconstruction and stage-level language supervision
introduced in Sec.~\ref{sec:spatiotemporal} for coordinated bimanual execution.

\textbf{Low-Coordination Bimanual Tasks.}
As shown in Fig.~\ref{fig:experiment2}(a), we evaluate
\textit{Bowl Placement}, \textit{Towel Storage}, and
\textit{Flower Arrangement}, covering rigid-object placement,
deformable-object manipulation, and precision placement, respectively.
The Single-Arm setting achieves success rates of \textbf{45\%},
\textbf{33\%}, and \textbf{42\%}, whereas RoboSTD improves them to
\textbf{86\%}, \textbf{62\%}, and \textbf{84\%}. Averaged across the
three tasks, the success rate increases from \textbf{40.0\%} to
\textbf{77.3\%}, while remaining close to the Oracle reference.
These results further demonstrate that RoboSTD-generated pseudo-bimanual
supervision can support real-world bimanual policy learning
across tasks with relatively limited inter-arm coordination requirements across diverse object types.

\textbf{High-Coordination Bimanual Tasks.}
We next consider \textit{Cup Collection} and \textit{Sandwich Making},
which require explicit spatial and temporal coordination between the two
arms. As shown in Fig.~\ref{fig:experiment2}(b),
\textit{RoboSTD w/o LLM} consistently improves over Single-Arm
supervision but remains below the full RoboSTD, indicating that sagittal-plane transfer and spatial rearrangement alone are insufficient for strongly coordinated bimanual behaviors. In \textit{Cup Collection}, the average
number of collected cups increases from \textbf{2.00} with Single-Arm
supervision to \textbf{2.76} with \textit{RoboSTD w/o LLM}, and further
to \textbf{4.22} with RoboSTD, slightly exceeding the Oracle result of
\textbf{3.92}. This highlights the benefit of arm assignment and conflict
constraints for organizing transferred skills across the shared
workspace. Similarly, in \textit{Sandwich Making}, the average completion
level improves from \textbf{55\%} to \textbf{63\%} and then to
\textbf{74\%}, compared with \textbf{72.5\%} for Oracle. The complete-task
success rate increases from \textbf{26\%} to \textbf{30\%} and finally
to \textbf{50\%}, matching the Oracle success rate of \textbf{50\%}.
These results demonstrate that LLM-guided spatio-temporal reconstruction
provides coordination benefits beyond geometric
Single-to-Dual transfer, particularly for shared workspaces and
long-horizon multi-stage tasks where arm assignment and
execution ordering are essential for coordinated bimanual behavior.

Overall, the real-world results reveal two complementary roles of
RoboSTD. Sagittal-plane mirroring and spatial rearrangement provide
transferable bimanual supervision, while LLM-guided spatio-temporal
reconstruction further organizes transferred skills through arm
assignment, precedence, and conflict constraints. On the
high-coordination tasks, RoboSTD slightly exceeds Oracle in average cup
collection (\textbf{4.22} vs.\ \textbf{3.92}) and sandwich completion
(\textbf{74\%} vs.\ \textbf{72.5\%}), while matching its task success
rate (\textbf{50\%}). These results validate RoboSTD for real-world Single-to-Dual transfer (\textbf{Q1})
and the role of structured spatio-temporal reconstruction in bimanual
coordination (\textbf{Q2}).

% ================================================================
\begin{figure}[!t]
\centering
\includegraphics[width=\columnwidth,keepaspectratio]
{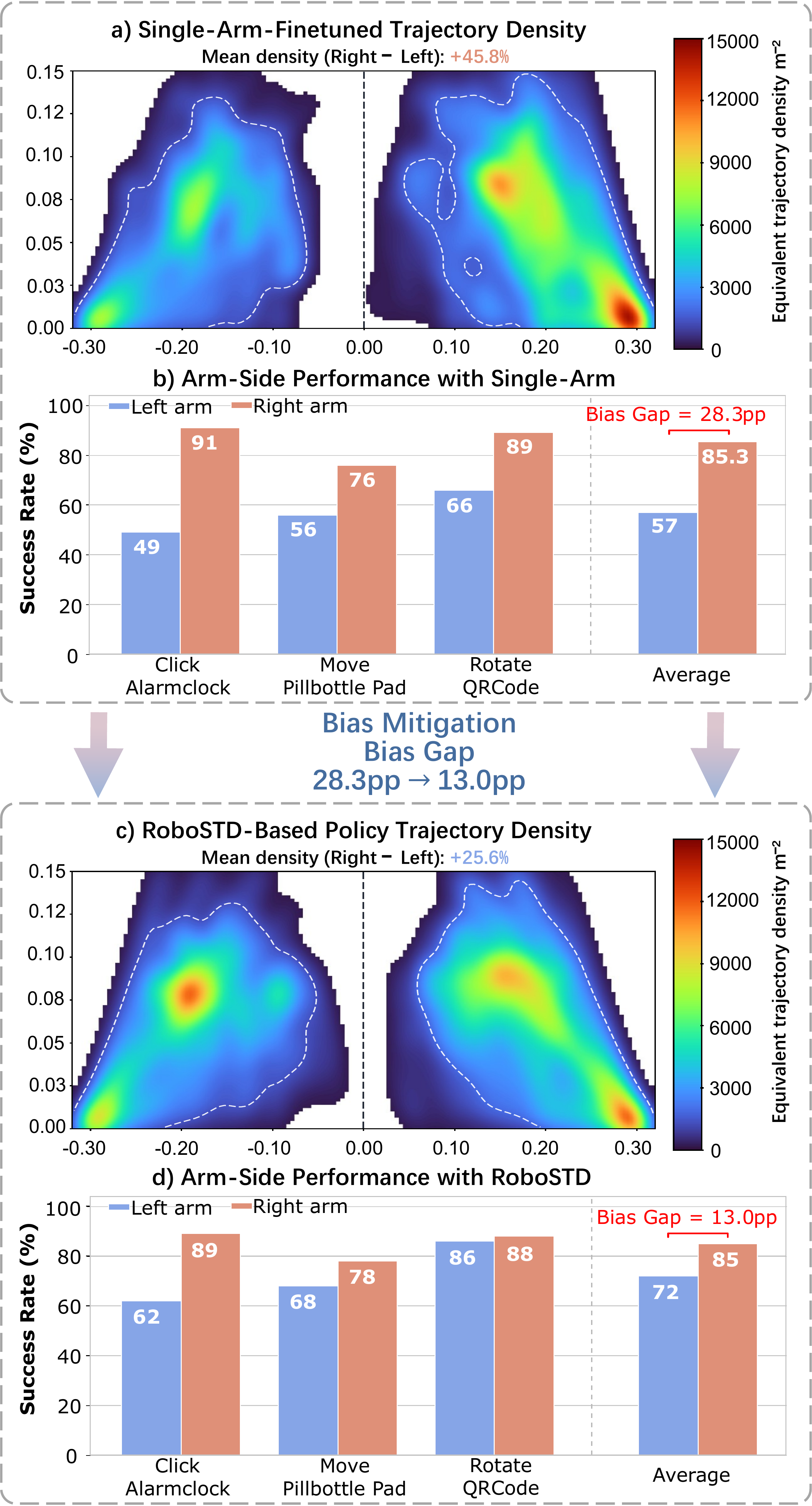}
\caption{\textbf{RoboTwin evaluation of arm-side spatial preference and
its mitigation with RoboSTD.}
(a--b) End-effector trajectory density and arm-side success rates of the
Single-Arm-finetuned baseline.
(c--d) Corresponding results of the RoboSTD-based policy on the three
selected RoboTwin tasks.}
\label{fig:experiment3}
\end{figure}

\subsection{Mitigating Arm-Side Spatial Preference}

To answer \textbf{Q3}, we evaluate whether RoboSTD can alleviate
right-side spatial preference in pretrained robot policies. We select
\textit{Click Alarmclock}, \textit{Move Pillbottle Pad}, and
\textit{Rotate QRCode}, the three RoboTwin tasks with the largest
right-to-left performance gaps in our preliminary 10-task evaluation.
For the Single-Arm baseline, we fine-tune pretrained $\pi_0$ on
\textbf{1,000} original single-arm demonstrations, with 100
demonstrations per task. For RoboSTD, the same 1,000 source
demonstrations are converted into pseudo-bimanual supervision to train
a RoboSTD-based model, which is then adapted to the three tasks
using their task-specific single-arm demonstrations. As shown in Fig.~\ref{fig:experiment3}, we compare the two training
settings from trajectory-distribution and arm-side performance
perspectives to examine whether RoboSTD improves workspace
balance and arm-side task performance.

As shown in Fig.~\ref{fig:experiment3}(a,c), the Single-Arm-finetuned
policy exhibits a clear right-side preference in its executed
trajectories, with a Density Gap of \textbf{45.8\%} on the
$\Delta X$--$\Delta Z$ workspace plane. In contrast, the RoboSTD-based
policy reduces the Density Gap to \textbf{25.6\%}, resulting in more
balanced left--right workspace coverage. The redistributed trajectory
density further shows that execution is no longer concentrated mainly
in the originally preferred right workspace, but extends more evenly
toward the underrepresented left side. This suggests that RoboSTD broadens the spatial utilization of transferred
skills across both arm workspaces, yielding more symmetric workspace
coverage and reducing the concentration of execution on the originally
preferred side.

A consistent trend is observed in task performance
(Fig.~\ref{fig:experiment3}(b,d)). Under Single-Arm supervision, the
average right- and left-arm success rates are \textbf{85.3\%} and
\textbf{57.0\%}, respectively, corresponding to a Bias Gap of
\textbf{28.3 pp}. With the RoboSTD-based policy, the average left-arm
success rate increases to \textbf{72.0\%}, while right-arm performance
remains nearly unchanged at \textbf{85.0\%}, reducing the Bias Gap to
\textbf{13.0 pp}. This reduction mainly results from improved left-arm
capability rather than degraded right-arm performance. Together with the
trajectory-density results, these findings show more balanced arm-side
behavior at execution and task-performance levels, supporting RoboSTD in
mitigating arm-side spatial preference in pretrained robot policies.

\section{Conclusion}
In this paper, we presented RoboSTD, a zero-shot Single-to-Dual data
construction framework for data-efficient bimanual robot learning.
RoboSTD constructs pseudo-bimanual supervision through sagittal-plane
mirroring and LLM-guided spatio-temporal reconstruction. The former
transfers single-arm skills to the contralateral workspace, while the
latter organizes original and mirrored skills through arm assignment,
precedence, and conflict constraints. Experiments in simulation and real-world environments demonstrate that RoboSTD can mitigate arm-side spatial preference in robot policies, provide effective pseudo-bimanual supervision, expand single-arm skills to broader bimanual workspaces, and improve long-horizon coordination. These results show that RoboSTD provides an effective way to reuse abundant single-arm demonstrations and alleviate the data bottleneck in bimanual robot learning. More broadly, our results suggest that geometric symmetry can serve as a practical bridge between unimanual and bimanual learning when combined with structured spatial and temporal supervision. This provides a promising direction for scaling bimanual robot learning without relying on large amounts of newly collected bimanual demonstrations for training.

% \section*{Acknowledgment}
% Add funding or acknowledgments here if needed.

\bibliographystyle{IEEEtran}
\bibliography{refs}

@inproceedings{zhao2023aloha,
  author    = {Tony Z. Zhao and Vikash Kumar and Sergey Levine and Chelsea Finn},
  title     = {{Learning Fine-Grained Bimanual Manipulation with Low-Cost Hardware}},
  booktitle = {Proceedings of Robotics: Science and Systems},
  year      = {2023},
  doi       = {10.15607/RSS.2023.XIX.016}
}

@inproceedings{chi2023diffusion,
  author    = {Cheng Chi and Siyuan Feng and Yilun Du and Zhenjia Xu and Eric Cousineau and Benjamin Burchfiel and Shuran Song},
  title     = {{Diffusion Policy: Visuomotor Policy Learning via Action Diffusion}},
  booktitle = {Proceedings of Robotics: Science and Systems},
  year      = {2023}
}

@inproceedings{black2024pi0,
  author    = {Kevin Black and Noah Brown and Danny Driess and Adnan Esmail and Michael Robert Equi and Chelsea Finn and Niccolo Fusai and Lachy Groom and Karol Hausman and Brian Ichter and Szymon Jakubczak and Tim Jones and Liyiming Ke and Sergey Levine and Adrian Li-Bell and Mohith Mothukuri and Suraj Nair and Karl Pertsch and Lucy Xiaoyang Shi and Laura Smith and James Tanner and Quan Vuong and Anna Walling and Haohuan Wang and Ury Zhilinsky},
  title     = {{$\pi_0$: A Vision-Language-Action Flow Model for General Robot Control}},
  booktitle = {Proceedings of Robotics: Science and Systems},
  year      = {2025},
  doi       = {10.15607/RSS.2025.XXI.010}
}

@misc{black2025pi05,
  author        = {{Physical Intelligence} and Kevin Black and Noah Brown and James Darpinian and Karan Dhabalia and Danny Driess and Adnan Esmail and Michael Equi and Chelsea Finn and Niccolo Fusai and Manuel Y. Galliker and Dibya Ghosh and Lachy Groom and Karol Hausman and Brian Ichter and Szymon Jakubczak and Tim Jones and Liyiming Ke and Devin LeBlanc and Sergey Levine and Adrian Li-Bell and Mohith Mothukuri and Suraj Nair and Karl Pertsch and Allen Z. Ren and Lucy Xiaoyang Shi and Laura Smith and Jost Tobias Springenberg and Kyle Stachowicz and James Tanner and Quan Vuong and Homer Walke and Anna Walling and Haohuan Wang and Lili Yu and Ury Zhilinsky},
  title         = {{$\pi_{0.5}$: A Vision-Language-Action Model with Open-World Generalization}},
  year          = {2025},
  eprint        = {2504.16054},
  archivePrefix = {arXiv},
  primaryClass  = {cs.LG},
  doi           = {10.48550/arXiv.2504.16054},
  note = {arXiv:2504.16054}
}

@inproceedings{kim2024openvla,
  author    = {Moo Jin Kim and Karl Pertsch and Siddharth Karamcheti and Ted Xiao and Ashwin Balakrishna and Suraj Nair and Rafael Rafailov and Ethan P. Foster and Pannag R. Sanketi and Quan Vuong and Thomas Kollar and Benjamin Burchfiel and Russ Tedrake and Dorsa Sadigh and Sergey Levine and Percy Liang and Chelsea Finn},
  title     = {{OpenVLA: An Open-Source Vision-Language-Action Model}},
  booktitle = {Proceedings of The 8th Conference on Robot Learning},
  series    = {Proceedings of Machine Learning Research},
  volume    = {270},
  pages     = {2679--2713},
  year      = {2025}
}

@inproceedings{ghosh2024octo,
  author    = {Dibya Ghosh and Homer Rich Walke and Karl Pertsch and Kevin Black and Oier Mees and Sudeep Dasari and Joey Hejna and Tobias Kreiman and Charles Xu and Jianlan Luo and You Liang Tan and Lawrence Yunliang Chen and Quan Vuong and Ted Xiao and Pannag R. Sanketi and Dorsa Sadigh and Chelsea Finn and Sergey Levine},
  title     = {{Octo: An Open-Source Generalist Robot Policy}},
  booktitle = {Proceedings of Robotics: Science and Systems},
  year      = {2024},
  doi       = {10.15607/RSS.2024.XX.090}
}

@inproceedings{oneill2023openx,
  author    = {{Open X-Embodiment Collaboration} and Abby O'Neill and Abdul Rehman and Abhiram Maddukuri and Abhishek Gupta and Abhishek Padalkar and Abraham Lee and Acorn Pooley and Agrim Gupta and Ajay Mandlekar and others},
  title     = {{Open X-Embodiment: Robotic Learning Datasets and RT-X Models}},
  booktitle = {Proceedings of the IEEE International Conference on Robotics and Automation},
  pages     = {6892--6903},
  year      = {2024},
  doi       = {10.1109/ICRA57147.2024.10611477}
}

@inproceedings{khazatsky2024droid,
  author    = {Alexander Khazatsky and Karl Pertsch and Suraj Nair and Ashwin Balakrishna and Sudeep Dasari and Siddharth Karamcheti and Soroush Nasiriany and Mohan Kumar Srirama and Lawrence Yunliang Chen and Kirsty Ellis and others},
  title     = {{DROID: A Large-Scale In-The-Wild Robot Manipulation Dataset}},
  booktitle = {Proceedings of Robotics: Science and Systems},
  year      = {2024},
  doi       = {10.15607/RSS.2024.XX.120}
}

@inproceedings{walke2023bridgedata,
  author    = {Homer Rich Walke and Kevin Black and Tony Z. Zhao and Quan Vuong and Chongyi Zheng and Philippe Hansen-Estruch and Andre Wang He and Vivek Myers and Moo Jin Kim and Max Du and Abraham Lee and Kuan Fang and Chelsea Finn and Sergey Levine},
  title     = {{BridgeData V2: A Dataset for Robot Learning at Scale}},
  booktitle = {Proceedings of The 7th Conference on Robot Learning},
  series    = {Proceedings of Machine Learning Research},
  volume    = {229},
  pages     = {1723--1736},
  year      = {2023}
}

@inproceedings{luo2024hajl,
  author    = {Shengcheng Luo and Quanquan Peng and Jun Lv and Kaiwen Hong and Katherine Rose Driggs-Campbell and Cewu Lu and Yong-Lu Li},
  title     = {{Human-Agent Joint Learning for Efficient Robot Manipulation Skill Acquisition}},
  booktitle = {Proceedings of the IEEE International Conference on Robotics and Automation},
  pages     = {1370--1377},
  year      = {2025},
  doi       = {10.1109/ICRA55743.2025.11127637}
}

@inproceedings{fu2024mobilealoha,
  author    = {Zipeng Fu and Tony Z. Zhao and Chelsea Finn},
  title     = {{Mobile ALOHA: Learning Bimanual Mobile Manipulation using Low-Cost Whole-Body Teleoperation}},
  booktitle = {Proceedings of The 8th Conference on Robot Learning},
  series    = {Proceedings of Machine Learning Research},
  volume    = {270},
  pages     = {4066--4083},
  year      = {2025}
}

@inproceedings{mandlekar2023mimicgen,
  author    = {Ajay Mandlekar and Soroush Nasiriany and Bowen Wen and Iretiayo Akinola and Yashraj Narang and Linxi Fan and Yuke Zhu and Dieter Fox},
  title     = {{MimicGen: A Data Generation System for Scalable Robot Learning using Human Demonstrations}},
  booktitle = {Proceedings of The 7th Conference on Robot Learning},
  year      = {2023}
}

@inproceedings{jiang2024dexmimicgen,
  author    = {Zhenyu Jiang and Yuqi Xie and Kevin Lin and Zhenjia Xu and Weikang Wan and Ajay Mandlekar and Linxi Fan and Yuke Zhu},
  title     = {{DexMimicGen: Automated Data Generation for Bimanual Dexterous Manipulation via Imitation Learning}},
  booktitle = {Proceedings of the IEEE International Conference on Robotics and Automation},
  year      = {2025},
  doi       = {10.1109/ICRA55743.2025.11127809}
}

@article{ordonez2024morphological,
  author  = {Daniel Felipe Ordo{\~n}ez Apraez and Giulio Turrisi and Vladimir Kostic and Mario Mart{\'i}n and Antonio Agudo and Francesc Moreno-Noguer and Massimiliano Pontil and Claudio Semini and Carlos Mastalli},
  title   = {{Morphological Symmetries in Robotics}},
  journal = {The International Journal of Robotics Research},
  volume  = {44},
  number  = {10--11},
  pages   = {1743--1766},
  year    = {2025},
  doi     = {10.1177/02783649241282422}
}

@inproceedings{li2025symdex,
  author    = {Zechu Li and Yufeng Jin and Daniel Ordonez-Apraez and Claudio Semini and Puze Liu and Georgia Chalvatzaki},
  title     = {{Morphologically Symmetric Reinforcement Learning for Ambidextrous Bimanual Manipulation}},
  booktitle = {Proceedings of The 9th Conference on Robot Learning},
  series    = {Proceedings of Machine Learning Research},
  volume    = {305},
  pages     = {1953--1974},
  year      = {2025}
}

@inproceedings{zhuang2025mirrorduo,
  author    = {Zheyu Zhuang and Ruiyu Wang and Giovanni Luca Marchetti and Florian T. Pokorny and Danica Kragic},
  title     = {{MirrorDuo: Reflection-Consistent Visuomotor Learning from Mirrored Demonstration Pairs}},
  booktitle = {Proceedings of The 9th Conference on Robot Learning},
  series    = {Proceedings of Machine Learning Research},
  volume    = {305},
  pages     = {2391--2408},
  year      = {2025}
}

@misc{chen2025robotwin2,
  author        = {Tianxing Chen and Zanxin Chen and Baijun Chen and Zijian Cai and Yibin Liu and Zixuan Li and Qiwei Liang and Xianliang Lin and Yiheng Ge and Zhenyu Gu and others},
  title         = {{RoboTwin 2.0: A Scalable Data Generator and Benchmark with Strong Domain Randomization for Robust Bimanual Robotic Manipulation}},
  year          = {2025},
  eprint        = {2506.18088},
  archivePrefix = {arXiv},
  primaryClass  = {cs.RO},
  note = {arXiv:2506.18088}
}

@misc{wu2025robocoin,
  author        = {Shihan Wu and Xuecheng Liu and Shaoxuan Xie and Pengwei Wang and Xinghang Li and Bowen Yang and Zhe Li and Kai Zhu and Hongyu Wu and Yiheng Liu and others},
  title         = {{RoboCOIN: An Open-Sourced Bimanual Robotic Data COllection for INtegrated Manipulation}},
  year          = {2025},
  eprint        = {2511.17441},
  archivePrefix = {arXiv},
  primaryClass  = {cs.RO},
  note = {arXiv:2511.17441}
}

@inproceedings{liu2025dcoda,
  author    = {I-Chun Arthur Liu and Jason Chen and Gaurav S. Sukhatme and Daniel Seita},
  title     = {{D-CODA: Diffusion for Coordinated Dual-Arm Data Augmentation}},
  booktitle = {Proceedings of The 9th Conference on Robot Learning},
  series    = {Proceedings of Machine Learning Research},
  volume    = {305},
  pages     = {3569--3588},
  year      = {2025}
}

@inproceedings{chen2025ropa,
  author    = {Jason Chen and I-Chun Arthur Liu and Gaurav S. Sukhatme and Daniel Seita},
  title     = {{ROPA: Synthetic Robot Pose Generation for RGB-D Bimanual Data Augmentation}},
  booktitle = {Proceedings of the IEEE International Conference on Robotics and Automation},
  year      = {2026}
}

@inproceedings{lu2024anybimanual,
  author    = {Guanxing Lu and Tengbo Yu and Haoyuan Deng and Season Si Chen and Yansong Tang and Ziwei Wang},
  title     = {{AnyBimanual: Transferring Unimanual Policy for General Bimanual Manipulation}},
  booktitle = {Proceedings of the IEEE/CVF International Conference on Computer Vision},
  pages     = {13662--13672},
  year      = {2025}
}

@misc{openai2025gpt41,
  author       = {{OpenAI}},
  title        = {Introducing {GPT-4.1} in the {API}},
  year         = {2025},
  month        = apr,
  howpublished = {\url{https://openai.com/index/gpt-4-1/}},
  note         = {Accessed: Sep. 13, 2026}
}

\clearpage
\appendices
\appendices

\section{Derivation of Mirror-Consistent Joint Mapping}
\label{app:joint_mapping}

This appendix provides a detailed derivation of the joint-space mirror
mapping used in RoboSTD and explains the origin of the platform-specific
sign matrix $\mathbf{\Lambda}$ introduced in the main text. For notational
simplicity, we use $\mathbf{q}^{L}$ in this appendix to denote the mirrored
target-arm configuration $\tilde{\mathbf{q}}^{L}$ in the main text.

Let the configuration space of an $n$-DoF robot arm be
\begin{equation}
\mathcal{Q} \subset \mathbb{R}^{n}.
\end{equation}
The joint state of the source arm, e.g., the right arm, is denoted as
\begin{equation}
\mathbf{q}^{R}
=
[q_1^R, q_2^R, \dots, q_n^R]^T,
\end{equation}
and the corresponding mirrored target-arm state, e.g., the left arm, is
denoted as
\begin{equation}
\mathbf{q}^{L}
=
[q_1^L, q_2^L, \dots, q_n^L]^T.
\end{equation}

We assume that corresponding joints of the two arms share consistent
actuation definitions and that, when both arms are at their zero
configurations, the overall kinematic structure is symmetric with respect
to the robot sagittal plane. Let the spatial reflection with respect to
the sagittal plane be represented by an orthogonal matrix
\begin{equation}
\mathbf{S} \in \mathbb{R}^{3 \times 3},
\end{equation}
where
\begin{equation}
\mathbf{S}
=
\mathbf{S}^{T}
=
\mathbf{S}^{-1},
\qquad
\det(\mathbf{S})=-1.
\end{equation}

For a point $\mathbf{p}$ in task space, its mirrored position is
\begin{equation}
\tilde{\mathbf{p}}
=
\mathbf{S}\mathbf{p}.
\end{equation}
For an end-effector pose
$\mathbf{T}=(\mathbf{R},\mathbf{p})$,
where $\mathbf{R}$ denotes the orientation and $\mathbf{p}$ denotes the
position, the mirrored pose is
\begin{equation}
\tilde{\mathbf{T}}
=
(\mathbf{S}\mathbf{R}\mathbf{S},
 \mathbf{S}\mathbf{p}).
\end{equation}
Thus, position and orientation are transformed consistently under the
same sagittal reflection.

The key issue in joint-space mapping is the reflection property of
rotational axes. Let
$\mathbf{R}(\mathbf{v},\theta)\in SO(3)$
denote a rotation of angle $\theta$ about a unit axis $\mathbf{v}$.
Under spatial reflection,
\begin{equation}
\mathbf{S}
\mathbf{R}(\mathbf{v},\theta)
\mathbf{S}
=
\mathbf{R}(\mathbf{S}\mathbf{v},-\theta).
\end{equation}
Hence, reflection changes both the rotation axis and the sign of the
rotation angle. Whether a mirrored joint angle preserves or reverses its
sign therefore depends on the geometric relationship between the target
joint axis and the reflected source joint axis.

\subsection{Physically Mirrored Arms}

We first consider the case where the left and right arms are physical
mirror copies of each other. In a common robot-centered frame, the
$i$-th joint axis of the left arm, $\mathbf{v}_i^L$, is the spatial
reflection of the corresponding right-arm axis $\mathbf{v}_i^R$:
\begin{equation}
\mathbf{v}_i^L
=
\mathbf{S}\mathbf{v}_i^R.
\end{equation}

To generate mirror-symmetric motion, the left-arm rotation must match
the reflected right-arm rotation:
\begin{equation}
\mathbf{R}(\mathbf{v}_i^L,q_i^L)
=
\mathbf{R}(\mathbf{S}\mathbf{v}_i^R,-q_i^R).
\end{equation}
Since
$\mathbf{v}_i^L=\mathbf{S}\mathbf{v}_i^R$,
the rotation axes are identical, and therefore
\begin{equation}
q_i^L
=
-q_i^R.
\end{equation}
Thus, for physically mirrored arms, the joint-space mapping becomes
\begin{equation}
\mathbf{q}^{L}
=
-\mathbf{q}^{R},
\end{equation}
which corresponds to the special case
$\mathbf{\Lambda}=-\mathbf{I}$
discussed in the main text.

\subsection{Pitch-, Yaw-, and Roll-like Joints in Isomorphic Arms}

A more common practical configuration uses two identical robot arms
mounted symmetrically. Such arms are not physical mirror copies, but
isomorphic manipulators with the same mechanical structure.

After transforming the corresponding joint axes into a common
robot-centered frame and accounting for the symmetric mounting
configuration, the axes of the two identical arms can be treated as
aligned:
\begin{equation}
\mathbf{v}_i^L
=
\mathbf{v}_i^R.
\end{equation}

The desired mirrored rotation of the source joint remains
\begin{equation}
\mathbf{R}
(\mathbf{S}\mathbf{v}_i^R,-q_i^R).
\end{equation}
Therefore, the target-arm joint angle $q_i^L$ must satisfy
\begin{equation}
\mathbf{R}(\mathbf{v}_i^R,q_i^L)
=
\mathbf{R}
(\mathbf{S}\mathbf{v}_i^R,-q_i^R).
\end{equation}

\begin{figure*}[!b]
\centering
\includegraphics[width=\textwidth]{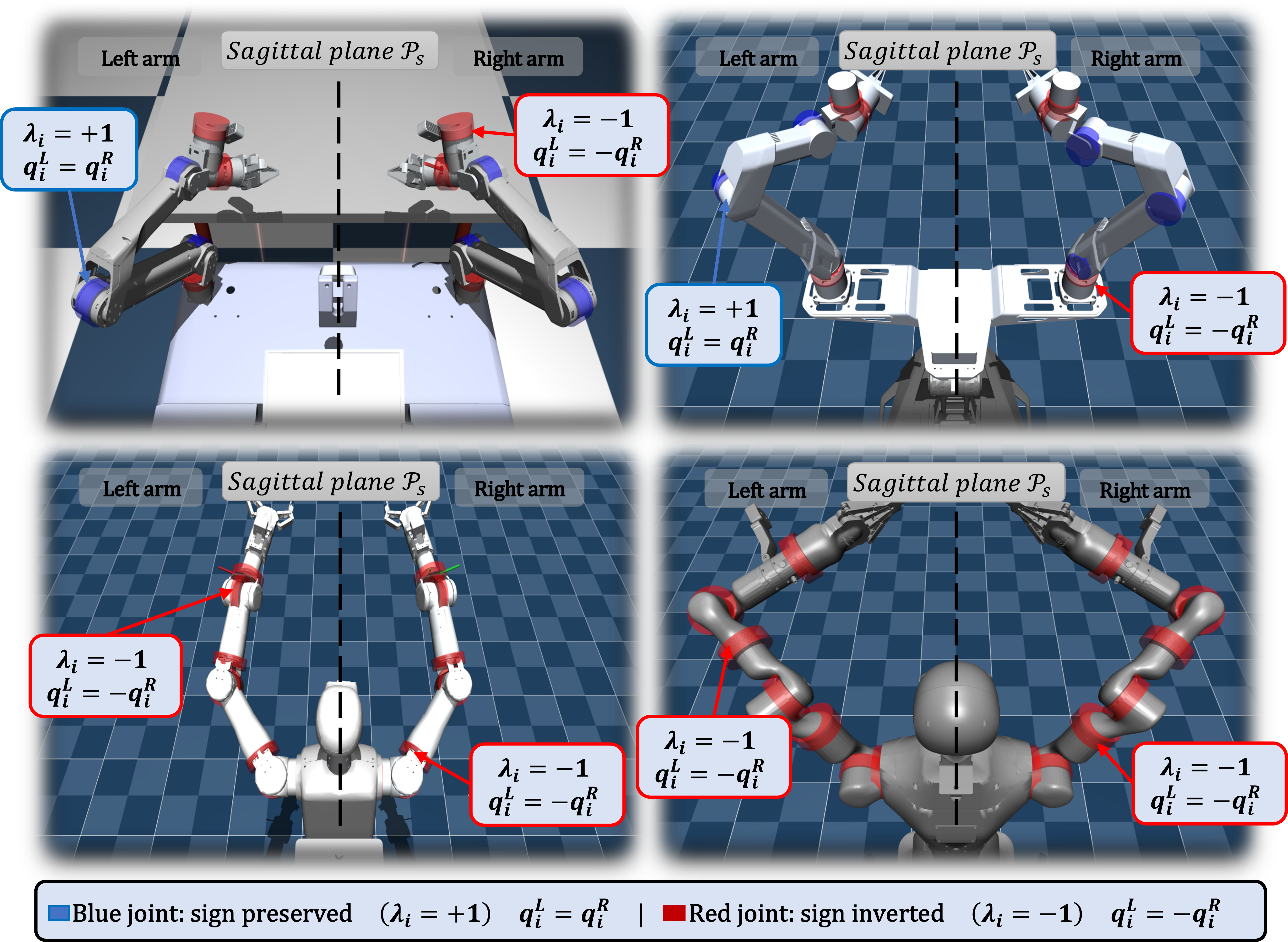}
\caption{
\textbf{Examples of mirror-consistent joint-sign assignments under sagittal reflection.}
For representative symmetrically mounted bimanual systems, the dashed line
indicates the sagittal plane $\mathcal{P}_s$. Blue joints denote sign-preserving mappings ($\lambda_i=+1$, $q_i^L=q_i^R$), whereas red joints denote sign-inverting mappings ($\lambda_i=-1$, $q_i^L=-q_i^R$). The same sign-assignment principle applies consistently across the illustrated robot morphologies. The assignment is determined by the reflected joint-axis relation and the robot mounting configuration, rather than by the joint name itself.}
\label{fig:joint_sign_examples}
\end{figure*}

The sign of $q_i^L$ is determined by how the joint axis
$\mathbf{v}_i^R$ transforms under sagittal reflection. For clarity,
assume that the sagittal plane is the $x$--$z$ plane of the
robot-centered frame and that the lateral direction is the $y$-axis.
The reflection reverses the $y$ direction while preserving directions
within the $x$--$z$ plane, leading to two representative cases.

\subsubsection{Pitch-like joints}

A pitch-like joint typically rotates about a lateral axis that is
approximately perpendicular to the sagittal plane. Its axis therefore
reverses under reflection:
\begin{equation}
\mathbf{S}\mathbf{v}_i^R
=
-\mathbf{v}_i^R.
\end{equation}
Substituting this relation into the mirrored rotation condition gives
\begin{equation}
\mathbf{R}(\mathbf{v}_i^R,q_i^L)
=
\mathbf{R}(-\mathbf{v}_i^R,-q_i^R).
\end{equation}
Using the axis-angle equivalence
\begin{equation}
\mathbf{R}(-\mathbf{v},-\theta)
=
\mathbf{R}(\mathbf{v},\theta),
\end{equation}
we obtain
\begin{equation}
\mathbf{R}(\mathbf{v}_i^R,q_i^L)
=
\mathbf{R}(\mathbf{v}_i^R,q_i^R),
\end{equation}
and hence
\begin{equation}
q_i^L
=
q_i^R.
\end{equation}
Therefore, pitch-like joints in such isomorphic bimanual systems
typically preserve their joint-angle sign under sagittal mirroring.

\subsubsection{Yaw- and roll-like joints}

Yaw-like and roll-like joints typically rotate about axes lying in, or
parallel to, the sagittal plane. Their axes are therefore preserved
under reflection:
\begin{equation}
\mathbf{S}\mathbf{v}_i^R
=
\mathbf{v}_i^R.
\end{equation}
Substituting this relation into the mirrored rotation condition gives
\begin{equation}
\mathbf{R}(\mathbf{v}_i^R,q_i^L)
=
\mathbf{R}(\mathbf{v}_i^R,-q_i^R),
\end{equation}
and thus
\begin{equation}
q_i^L
=
-q_i^R.
\end{equation}
Therefore, yaw-like and roll-like joints in such isomorphic bimanual
systems typically reverse their joint-angle sign under sagittal
mirroring.

\subsection{Unified Representation}

For an isomorphic bimanual system composed of two identical robot arms,
the joint-space mirror mapping can be written in the unified form
\begin{equation}
\mathbf{q}^{L}
=
\mathbf{\Lambda}\mathbf{q}^{R},
\end{equation}
where
\begin{equation}
\mathbf{\Lambda}
=
\mathrm{diag}(\lambda_1,\lambda_2,\dots,\lambda_n),
\qquad
\lambda_i\in\{+1,-1\}.
\end{equation}

The diagonal element $\lambda_i$ is determined by the relationship
between the joint axis and its sagittal reflection:
\begin{equation}
\lambda_i
=
\begin{cases}
+1, & \mathbf{S}\mathbf{v}_i^R=-\mathbf{v}_i^R,\\
-1, & \mathbf{S}\mathbf{v}_i^R=\mathbf{v}_i^R.
\end{cases}
\label{eq:appendix_sign_rule}
\end{equation}

Figure~\ref{fig:joint_sign_examples} provides representative examples of
this sign-assignment rule across different symmetric bimanual morphologies.
Although the joint layouts and viewing configurations vary, the same
geometric criterion applies: joints whose axes reverse under sagittal
reflection preserve the joint-angle sign, whereas joints whose axes remain
aligned under reflection invert the sign.

Equation~\eqref{eq:appendix_sign_rule} is the isomorphic-arm
specialization of the general cross-arm sign rule in
Eq.~\eqref{eq:joint_sign} of the main text, obtained after expressing
the corresponding joint axes in a common robot-centered frame.

For common isomorphic bimanual configurations, this relation can be
interpreted as
\begin{equation}
\lambda_i
=
\begin{cases}
+1, & \text{pitch-like joint},\\
-1, & \text{yaw-like or roll-like joint}.
\end{cases}
\end{equation}

The pitch, yaw, and roll terminology depends on the robot coordinate
frame and joint-axis definitions. RoboSTD therefore does not determine
$\mathbf{\Lambda}$ from joint names. Instead, the sign matrix is derived
from the joint-axis directions specified by the robot URDF together with
the robot mounting configuration. Once determined,
$\mathbf{\Lambda}$ is fixed for the corresponding robot platform and
applied consistently to mirrored proprioceptive states and joint-space
actions. In particular,
\begin{equation}
\tilde{\mathbf{q}}^{L}
=
\mathbf{\Lambda}\mathbf{q}^{R},
\qquad
\tilde{\mathbf{a}}^{L}
=
\mathbf{\Lambda}\mathbf{a}^{R},
\end{equation}
with the same platform-specific sign convention used throughout RoboSTD
data construction.

\end{document}